\documentclass[runningheads]{llncs}
\usepackage{graphicx}

\usepackage{tikz}
\usepackage{comment}
\usepackage{amsmath,amssymb} 
\usepackage{graphicx}
\usepackage{booktabs}
\usepackage{multirow}
\usepackage{subcaption}
\usepackage{enumitem}
\usepackage{color}

\usepackage{pifont}
\newcommand{\cmark}{\ding{51}}%
\newcommand{\xmark}{\ding{55}}%

\usepackage[pagebackref=true,breaklinks=true,letterpaper=true,colorlinks,bookmarks=false,linkcolor=blue]{hyperref}
\usepackage[width=122mm,left=12mm,paperwidth=146mm,height=193mm,top=12mm,paperheight=217mm]{geometry}
\newcommand{\first}[1]{{\textcolor{red}{\textbf{#1}}}}
\newcommand{\second}[1]{{\textcolor{blue}{\textbf{#1}}}}

\begin{document}
\pagestyle{headings}
\mainmatter
\def\ECCVSubNumber{5217}  

\title{FindIt: Generalized Localization with Natural Language Queries} 

%
%
\author{Weicheng Kuo,
Fred Bertsch,
Wei Li,
AJ Piergiovanni,
Mohammad Saffar,
Anelia Angelova}
%
\institute{Google Research, Brain Team \\ \email{\{weicheng,fredbertsch,mweili,ajpiergi,msaffar,anelia\}@google.com}}
\authorrunning{W. Kuo et al.}

\maketitle

\begin{abstract}
We propose FindIt, a simple and versatile framework that unifies a variety of visual grounding and localization tasks including referring expression comprehension, text-based localization, and object detection. Key to our architecture is an efficient multi-scale fusion module that unifies the disparate localization requirements across the tasks. In addition, we discover that a standard object detector is surprisingly effective in unifying these tasks without a need for task-specific design, losses, or pre-computed detections. Our end-to-end trainable framework responds flexibly and accurately to a wide range of referring expression, localization or detection queries for zero, one, or multiple objects. Jointly trained on these tasks, FindIt outperforms the state of the art on both referring expression and text-based localization, and shows competitive performance on object detection. Finally, FindIt generalizes better to out-of-distribution data and novel categories compared to strong single-task baselines. All of these are accomplished by a single, unified and efficient model. The code will be released.\footnote{Please see the project page: \href{https://sites.google.com/view/findit-eccv22/home}{https://sites.google.com/view/findit-eccv22/home}.}
\end{abstract}

\section{Introduction}
\label{sec:intro}

Natural language enables flexible descriptive queries about images. The interaction between text queries and images grounds linguistic meaning in the visual world, facilitating a stronger understanding of object relationships, human intentions towards objects, and interactions with the environment. The research community has studied visual grounding through tasks including phase grounding, object retrieval and localization, language-driven instance segmentation, and others ~\cite{plummer2017FlickrEntities,Referitgame,plummer2020revisiting,hu2016natural,tuay2018dynamic,wang2016structured,hu2016segmentation,gupta2020contrastrive}.

\begin{figure*}[t]
	\includegraphics[width=0.98\linewidth]{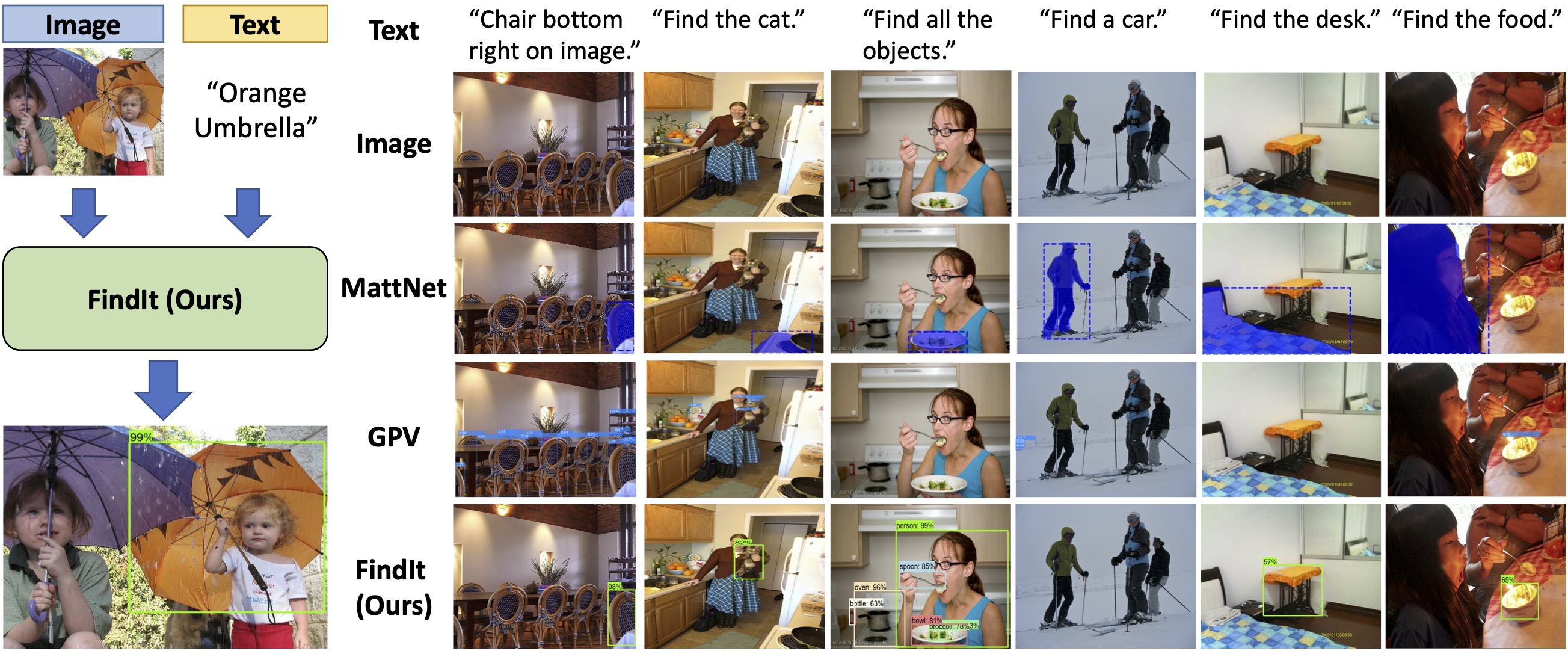}
	\caption{
	FindIt is a general-purpose model for visual grounding and localization tasks (left). The input is an image-text pair specifying the objects of interest using natural language, and the outputs are a set of bounding boxes and classification scores. Specifically, FindIt addresses
	the following tasks (col. 1-3): referring expression comprehension (col. 1), text-based localization (col. 2), and the object detection task by an optional generic prompt e.g. ``Find all the objects.'', (col. 3). Furthermore, FindIt can respond accurately when the referred object is absent (col. 4), or when it is tested on out-of-distribution (OOD) images and with novel category names, e.g. ``desk'', where ``dining table'' is the closest category in the training set (col. 5). FindIt can also locate objects, referred to by novel super-category names e.g. ``food'' (col. 6). We compare to MattNet~\cite{yu2018mattnet} and GPV~\cite{gupta2021gpv} in all these scenarios. (Best viewed in color)}
	\label{fig:teaser}
\end{figure*}

Among the most popular visual grounding tasks is referring expression comprehension (REC), which localizes an object given a referring text~\cite{yu2016modeling,mao2016generation,Referitgame}. This task often requires complex reasoning on prominent objects. A highly related semantic localization task is object detection (DET), which seeks to detect all objects from a predefined set of classes without text inputs~\cite{Papageorgiou1998ageneral,rowley1995human,vaillant1994an,Viola01robustreal-time,Girshick2014rich,redmon2016yolo}. In contrast to REC, this task requires the accurate classification and localization of small, occluded objects. At the intersection of the two is text-based localization~\cite{gupta2021gpv,hinami2018discriminative} (LOC), in which a simple category-based text query prompts the model to detect the objects of interest.

Due to their highly dissimilar task properties, REC, DET, and LOC are mostly studied through separate benchmarks with most models only dedicated to one task \cite{yang2019fast,faster,gupta2021gpv}. As a result, existing models have not adequately synthesized information from the three tasks to achieve a more holistic visual and linguistic understanding. REC models, for instance, are trained to predict one object per image, and often struggle to localize multiple objects\footnote{Technically, many REC models can localize more than one object, but they often struggle  because they are only trained to predict one object per image on REC data.}, reject negative queries, or detect novel categories (see ~\autoref{fig:teaser}). In addition, DET models are unable to process text inputs, and LOC models often struggle to process complex queries such as “Chair bottom right on image” (see ~\autoref{fig:teaser}). Lastly, none of the models can generalize sufficiently well beyond the their training data and categories.

To address these limitations, we propose a unified visual localization approach which we call FindIt. Key to our architecture is a multi-level cross-modality fusion module which can perform complex reasoning for REC and simultaneously recognize small and challenging objects for LOC and DET. 
To unify the disparate demands of these tasks, the module efficiently fuses and learns features across many levels of abstraction.
Concretely, we utilize the more expressive cross-attention fusion on lower resolution features, and the more efficient product fusion on higher resolution features to combine the best of both worlds.
Last but not least, we discover that a standard object detector and detection losses \cite{faster} are sufficient and surprisingly effective for REC, LOC, and DET tasks without a need for task-specific design and losses \cite{deng2021transvg,gupta2021gpv,liu2019improving,Luo2020MultiTaskCN,yang2020improving,yang2019fast,yu2018mattnet}. In short, FindIt is a simple, efficient, and end-to-end trainable model for unified visual grounding and object detection.

By learning REC, LOC, and DET jointly in one model, FindIt acquires a more holistic and versatile capability for visual grounding than its single-task counterparts. Notably, FindIt surpasses the state of the art on REC and LOC, and demonstrates competitive performance on DET. Moreover, unlike existing task-specific models, FindIt accomplishes these in a single model that can respond flexibly to a wide range of referring expression and localization queries, solve the standard detection task, and generalize better to novel data and classes. In summary, our contributions are:

\begin{itemize}[leftmargin=*]
\setlength\itemsep{0.00em}

    \item We propose FindIt, a simple and versatile framework for visual grounding and detection tasks. In contrast to task-specific models, a single FindIt model can respond flexibly to a wide range of referring expression and localization queries, solve the standard detection task, and generalize better to novel data and classes. 

    \item We propose an efficient multi-scale cross-attention fusion module to unify the disparate task requirements between REC, LOC, and DET. Using the fused features, we discover that a standard detector and detection losses are surprisingly effective for all tasks without a need for task-specific design or losses.
    
    \item We surpass the state of the art on REC and LOC, and show competitive DET performance within a single, unified and efficient model.
\end{itemize}

\vspace{-3mm}
\section{Related Work}
\vspace{-2mm}

\textbf{Referring Expression Comprehension (REC)} and phrase grounding tasks ~\cite{mao2016generation,Nagaraja2016modeling,yu2016modeling,yu2018mattnet,qiao2020referring,yang2019fast,mdetr,yang2020improving,plummer2017FlickrEntities} require the models to ground linguistic elements in the image. Several datasets which enable and enrich the study of these tasks have been proposed ~\cite{yu2016modeling,mao2016generation,hudson2019gqa,clevr2017clevr,Referitgame,chen2020copsref,plummer2017FlickrEntities}. Yu et al. and Mao et al.~\cite{yu2016modeling,mao2016generation} expand the COCO benchmark with referring expression annotations, while the Referit game~\cite{Referitgame} crowd-sources such labels through game-play. One-stage ~\cite{chen2018real,liao2020real,yang2019fast,deng2021transvg,Luo2020MultiTaskCN} and two-stage~\cite{zhang2018grounding,yu2018mattnet,hu2017modeling,wang2019neighbourhood,yang2019dynamic,liu2019learning} methods have been popular for these tasks. \textbf{Object Detection (DET)} task is well established and has a plethora of approaches ~\cite{Girshick2014rich,redmon2016yolo,faster,he2017MaskRCNN,lin2017focal} and benchmarks~\cite{mscoco,Objects365}. The goal is to identify the bounding boxes of a set of pre-defined classes without prompting by text. Many recent approaches have started to study the open-set and zero-shot settings ~\cite{djamija2020overlooked,bansal2018zeroshot,xian2018zeroshotlearning,zhu2018zsdetection,vild-zero-shot}. \textbf{Text-based Localization (LOC)} has been recently proposed alongside other vision and language tasks~\cite{gupta2021gpv,hinami2018discriminative}.  
Text-based localization is similar to the referring expression comprehension task. The text query specifies an object class to be localized. This task is typically derived from standard detection datasets~\cite{mscoco,krishnavisualgenome}. Early results with this tasks are presented by~\cite{gupta2021gpv,hinami2018discriminative} where the focus has been on a single object ~\cite{hinami2018discriminative}. FindIt extends this capability to localize multiple objects of any given category or detect all objects of a given vocabulary through a free-form text prompt.

\noindent \textbf{Multi-modal Vision-Language Learning.}
Large amounts of vision and language work are present, such as visual-grounding ~\cite{gan2017vqs,wang2018learning,liu2017recurrent,zhao2018weakly,yu2018mattnet,plummer2020revisiting}, image captioning~\cite{chen2015cococaptions,anderson2018bottomup,beer2021cc12m}, visual question and answering (VQA)~\cite{jiang2020in,das2017visual}, visual reasoning~\cite{sihr2017acorpus,zellers2019vcr,snli-ve}, image-text retrieval~\cite{align,clip,srinivasan2021wit}, and video-text learning ~\cite{huang2020multimodal,lin2021vx2text,xu2016mstvtt,das2017visual}. Many approaches to vision-language learning leverage large-scale image-text pre-training or pre-computed detections ~\cite{vilbert2020,visualbert,tan2019lxmert,chen2020uniter,li2020oscar,huang2021seeing,xu2021e3elvp,zhou2020unifiedVLP,kim2021vilt,ernievil,virtex,loctex,beer2021cc12m,clip}. In particular, many methods underscore the importance of localization to increase the success of related vision-and-language understanding/reasoning tasks such as VQA and CLEVR \cite{chen2020uniter,li2020oscar,vinvl,villa,anderson2018bottomup,changpinyo2021telling,mdetr}.

\noindent \textbf{Vision and Language Feature Fusion.}
Recently, the Transformer~\cite{vaswani2017attention} and its cross-modality variants~\cite{vilbert2020,chen2020uniter,kant2021contrast} have been popular fusion choices for vision-language tasks.
To localize objects at various scales, existing REC works have used multi-level fusion by applying activation and product fusion \cite{Luo2020MultiTaskCN,yang2020improving} or concatenation and convolution fusion \cite{yang2019fast}. Inspired by recent works \cite{deng2021transvg,chen2020uniter,vilbert2020,kant2021contrast} on single-scale cross-attention, we propose multi-scale fusion to satisfy the disparate requirements of REC and detection tasks, where REC requires complex reasoning while detection requires accurate localization and recognition. The fusion module enables us to unify these tasks in a single model and surpass the state of the art on REC, LOC and maintains competitive DET performance.

\noindent \textbf{Multi-task Learning for Visual Grounding and Object Localization.}
Existing approaches have combined grounding and localization tasks with text-generation tasks such as VQA, captioning, visual entailment~\cite{lu202112in1,cho2021unifying,hu2021unit}, and have leveraged pretraining or joint training with similar localization tasks~\cite{Luo2020MultiTaskCN,li2021grounded,mdetr}. Hu et al.~\cite{hu2021unit} unifies a detection task with text-generation tasks. GPV~\cite{gupta2021gpv} combines text-based localization with VQA by generating both boxes and text for each input image/text pairs. MCN \cite{Luo2020MultiTaskCN} jointly learn REC and RES (Referring Expression Segmentation) to show the benefits of multi-task learning for both. GLIP \cite{li2021grounded} formulates object detection as phrase grounding and combines detection, caption, and grounding datasets for zero/few-shot detection. M-DETR~\cite{mdetr} uses many grounding datasets in a phrase grounding pretraining. Similar to MCN, FindIt unifies semantically similar tasks to study the benefits of multitask learning. Different from M-DETR and GLIP, FindIt uses only COCO and RefCOCO data without a need for pretraining on external data.

\vspace{-2mm}
\section{Method}
\vspace{-2mm}
\subsection{Overview}

The goal of FindIt is to unify a family of semantically-related localization tasks: 1) referring expression comprehension (REC), 2) text-based localization (LOC), and 3) detection (DET). To accomplish this, FindIt produces a set of boxes/classes when given an RGB image and a text query (see \autoref{tab:task_comparison}). The architecture (Section \ref{sec:arch}) includes an image encoder, a text encoder, a fusion model, and a set of box/class prediction heads. The fusion model (Section \ref{sec:fusion}) takes multi-scale features from the image encoder and fuses them with the text encoder features. The box/class heads take the fused features as input and produce a set of bounding boxes, their categories and confidence scores. All tasks share the same architecture, losses, and weights.

\begin{table}[t]
    \centering
    \caption{FindIt tasks comparison. FindIt unifies the referring expression (REC), text-based localization (LOC) and detection (DET) tasks.}
    \label{tab:task_comparison}
    \scalebox{0.68}{  
    \begin{tabular}{|c|c|c|c|c|c|c|}
    \toprule
    Task & Text Input & Output & Image Size & Loss and Architecture & Metric \\
    \midrule
    REC & Expr. for one object & One box & 256 \cite{yang2019fast,peng2020largescale} / 640\cite{deng2021transvg} & Ref-specific or DETR loss/arch. \cite{yang2019fast,peng2020largescale,deng2021transvg} & Precision \\
    LOC & Expr. for one class & Many boxes & 640 \cite{gupta2021gpv} & DETR loss and DETR + image-text fusion \cite{gupta2021gpv} & AP50 \\
    DET & None / Task prompt & Many boxes/classes & 1333 \cite{he2017MaskRCNN,Detectron2018} & Two-stage \cite{faster}, one-stage \cite{redmon2016yolo,lin2017focal}, transformer \cite{detr} & AP \\
    \midrule
    FindIt & All the above & Many boxes/classes & 640 & Two-stage detector loss \cite{faster} + image-text fusion & All \\
    \bottomrule
    \end{tabular}
    }
\end{table}

\vspace{-2mm}
\subsection{Task Definitions}
\label{sec:tasks}

\autoref{tab:task_comparison} shows a comparison of the FindIt sub-tasks. Since these tasks are similar in nature, our goal is to unify, and consider them jointly. We define them as follows:
\begin{itemize} [leftmargin=*]
\setlength\itemsep{0.00em}
    \item \textbf{REC}: In the referring expression comprehension task, inputs are an image and a user query about a specific (often prominent) object in the image. The expected output is one bounding box around the correct object. While natural queries may invoke multiple objects, this task is limited to providing a single box as an answer. We adopt the standard precision@1 metric.
    \item \textbf{LOC}: In the text-based localization task, inputs are an image and a query about a category, e.g. ``Find the cars'' \cite{gupta2021gpv}. The expected output is a set of bounding boxes around all objects in that category. This task challenges the model to only predict the relevant objects based on the query. We follow the AP50 metric proposed by \cite{gupta2021gpv}.
    \item \textbf{DET}: In the detection task, inputs are an image and a standard query, ``Find all the objects''. The expected outputs are bounding boxes around the objects of categories present in the dataset and their classes, but as we show in \autoref{tab:gen}, FindIt can generalize to novel categories via text-based localization. Our modification allows us to share the same vision and language interface with the other tasks. We adopt the standard mAP metric in detection \cite{mscoco}. 
\end{itemize}

\begin{figure*}[t]
    \centering
    \includegraphics[width=0.43\linewidth]{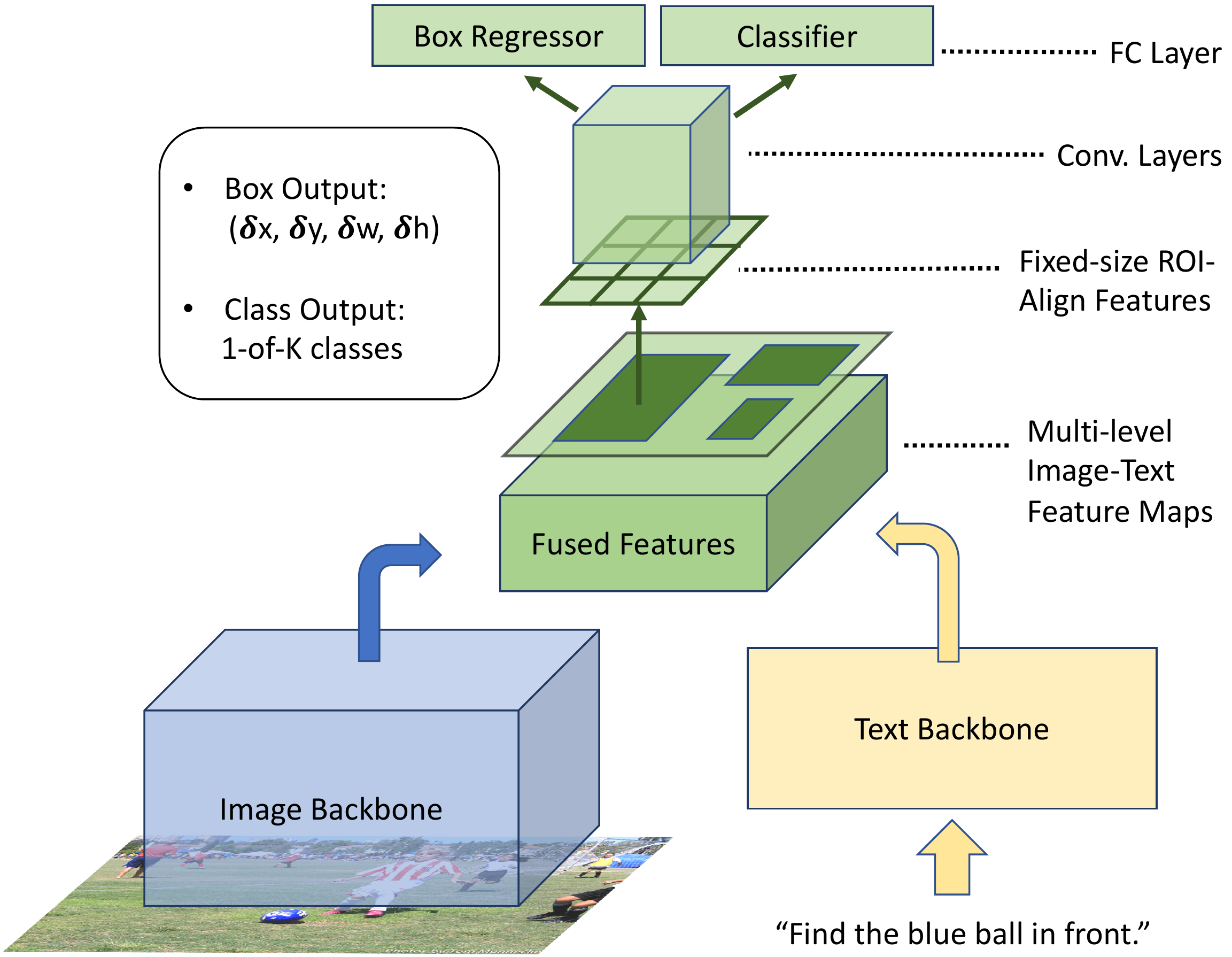}
    \includegraphics[width=0.53\linewidth]{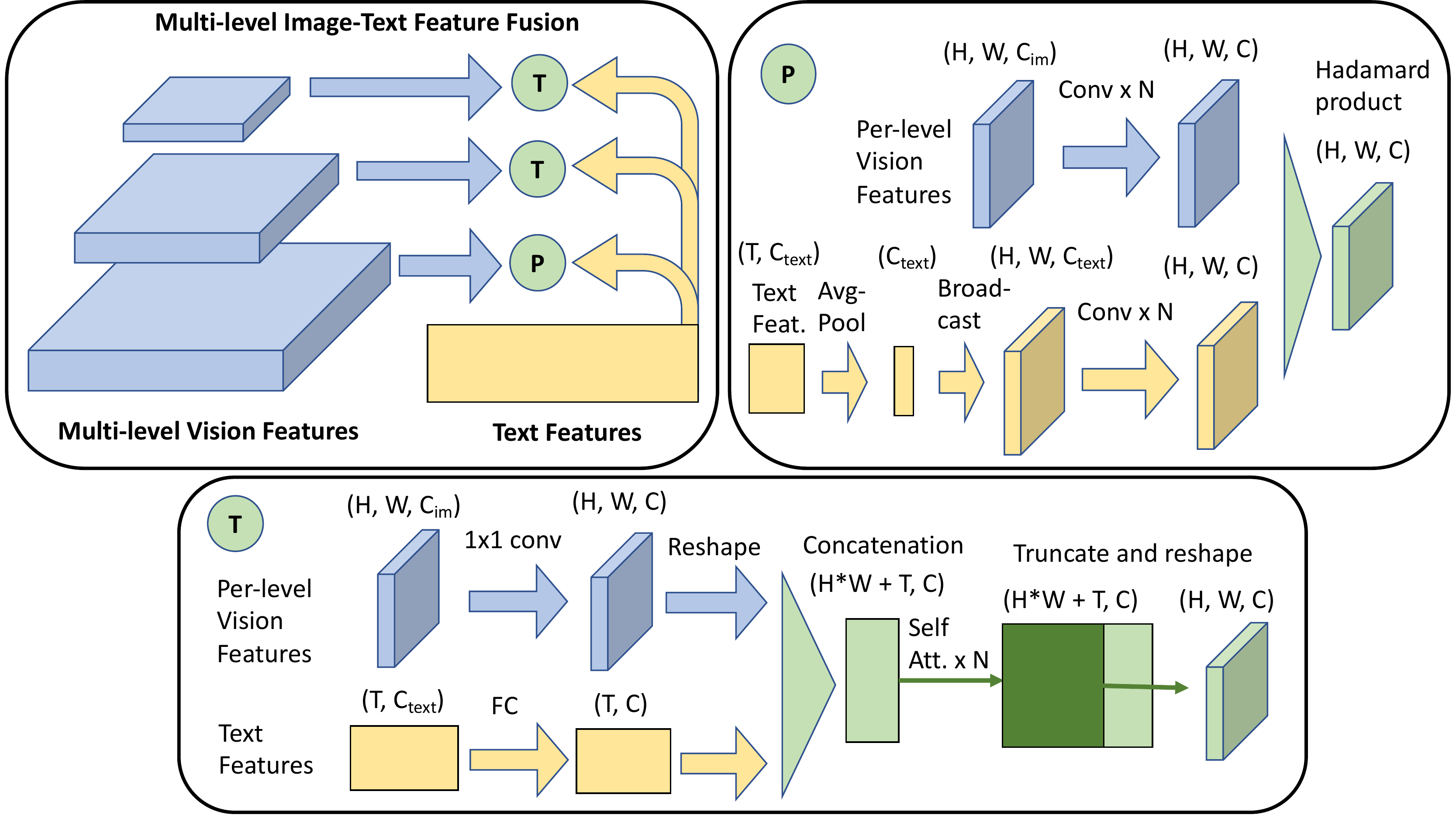}
    \caption{(Left) Our main architecture accepts an image and a query text as inputs, and processes them separately in image/text backbones before applying the multi-level fusion. We feed the fused features to region proposal network to generate candidate regions and then extract the region features for box regression and classification. (Right) Our multi-level image-text fusion module (top-left) uses transformer fusion blocks (T), and product fusion blocks (P) at the higher/lower levels of the feature maps respectively.}
    \label{fig:arch}
    \vspace{-2mm}
\end{figure*}

\vspace{-2mm}
\subsection{Network Architecture}
\label{sec:arch}
Our network architecture is simple and extensible: it includes an image encoder, a text encoder, a fusion model, and box/class prediction heads (\autoref{fig:arch}). All parameters are shared by all  tasks, i.e. there are \textit{no task-specific parameters}. The image encoder is a ResNet backbone which yields multi-level features. The text encoder is a T5 transformer~\cite{T5} model which encodes a query sentence as a series of token features. The fusion model fuses the multi-level image features with token features (Section~\ref{sec:fusion}). 
We fuse the image and text features at the image level, as it allows more flexibility to adapt visual representation to various queries. After the fusion, we apply the standard region proposer \cite{faster} and box/class decoders \cite{faster}. Our design can tackle any task that predicts multiple objects and their classes given an image and a text query (optional). Although we use FRCNN \cite{faster} in this work, our unification approach is agnostic to the choice of detectors and other detectors are also viable \cite{detr,lin2017focal,redmon2016yolo}\footnote{The detector head may also be adapted from existing visual grounding models such as~\cite{deng2021transvg,yang2019fast}, but we leave this for future studies.}.

\vspace{-2mm}
\subsection{Multi-level Image-Text Fusion}
\label{sec:fusion}
To combine these different localization tasks, one major challenge is that they are created around different domains and with different goals (see \autoref{tab:task_comparison}). For example, the referring expression task primarily references prominent objects in the image rather than small, occluded or faraway objects such that low resolution images would suffice. In contrast, the detection task aims to detect objects with various sizes and occlusion levels in higher resolution images. Apart from these benchmarks, the general visual grounding problem is inherently multiscale, as natural queries can refer to objects of any size. This motivates our multi-level image-text fusion model for efficient processing of higher resolution images over different localization tasks. 

We fuse multi-level image features with the text features using a Transformer-based cross attention module~\cite{vaswani2017attention} (See \autoref{fig:arch}). The vision features at each level are fused with the text features. A feature pyramid \cite{lin2017feature} fuses features across resolutions by progressively up-sampling the higher level fused features to the resolution of lower level features.

The transformer fusion works as follows (see bottom right of \autoref{fig:arch}). We first use a linear layer to project the vision and text features into the same dimension at each level. 
Next, we collapse the spatial dimension of vision features into a sequence and concatenate it with the text sequence features. 
We compute the relative position bias based on the total length of the concatenated sequence before applying the self-attention layers. 
As self-attention is intractable with large feature maps, we apply product fusion (see top right of \autoref{fig:arch}) for the early high resolution feature maps (i.e. F2 and F3), and use self-attention for the smaller, higher level feature maps (i.e. F4 and F5). Ablation studies show the benefits of multi-level fusion and self-attention for handling complex queries (see Section~\ref{sec:ablation}). Finally, we truncate and reshape the fused features to the same spatial dimensions as the input vision features.

\subsection{Task Unification and Multi-task Learning}
The three localization tasks must be unified in terms of model, loss, and inputs so they can be trained together. The implications of unification are significant. First, all tasks can share the same model during both training and inference time. Second, the unification of inputs and loss enables us to efficiently train on multiple datasets. Lastly, the model can leverage information from other tasks, which allows the transfer of visual concepts and enables zero-shot applications. For example, we can learn long-tail concepts from the referring expression task and transfer them to other localization tasks. 

Apart from the unified architecture (see Section~\ref{sec:arch} and \ref{sec:fusion}), datasets are adapted to the different tasks as follows. For the localization task, detection datasets are adapted by generating a set of queries over the categories present in the image. For any present category, the text query takes the form ``Find the X'' where X is the category name. The objects corresponding to that category are labeled as foreground and the other objects as background. At training time, we randomly sample a text query and corresponding objects from each image. For the detection task, detection datasets are adapted by adding a static task prompt such as ``Find all the objects''. We found that the specific choice of prompts are not important for LOC and DET tasks (see~\autoref{tab:text_prompt}).

After adaptation, all tasks in consideration share the same inputs and outputs\textemdash an image input, a text query, and a set of output bounding boxes/classes. We then combine the datasets and train on the mixture. At training time, we use a mixing ratio of 1:1:1 between DET:LOC:REC tasks in each minibatch. 
To ensure each dataset is sampled adequately, we use a larger batch size of 256 split among the 3 tasks.
To make the image size uniform across tasks (see \autoref{tab:task_comparison}), we adopt the LOC task’s image size of 640~\cite{gupta2021gpv} as a middle ground. This is larger but comparable to the image size of REC task~\cite{yang2019fast,yang2020improving,deng2021transvg}. It is smaller than the size of DET task’s images~\cite{he2017MaskRCNN} which might limit performance on smaller objects.

Finally, we unify the losses of all tasks. The losses we use are box classification and regression loss, region proposal classification and regression loss, and weight decay. The loss formulation and relative weights follow~\cite{faster} without any task-specific modification. All losses have equal weights across tasks. We note that it is unclear how to use existing grounding models out-of-the-box for task unification due to the task-specific architectures, losses, and training strategies \cite{deng2021transvg,gupta2021gpv,liu2019improving,Luo2020MultiTaskCN,yang2020improving,yang2019fast,yu2018mattnet}.

\subsection{Implementation details}
FindIt uses a region proposer (RPN) \cite{faster}, class predictor \cite{faster} and a class-agnostic box regressor \cite{faster} shared among all tasks. The class decoder has the same number of outputs as the detection vocabulary size (i.e. 80 for COCO), as it primarily serves the detection task.
We note that no pre-computed detections are used in FindIt as in many two-stage referring expression models \cite{yu2018mattnet,liu2019improving}.

FindIt image encoder is initialized from the ResNet50 model pretrained on COCO detection. FindIt text encoder is initialized from T5-base \cite{T5} pretrained checkpoint. All other modules are trained from scratch, including the multi-level fusion model, feature pyramid network \cite{lin2017feature}, the region proposal network (RPN) and the box/class decoders \cite{faster}. All hyper-parameters of the feature pyramid, RPN and box/class decoder heads follow the Faster R-CNN \cite{Detectron2018}.

We set the batch size to 256 split among 3 tasks DET:LOC:REC with mixing ratio 1:1:1 in the minibatch. The ratio was chosen for simplicity and has room for further optimization. We train the model for 150k steps on a learning rate of 0.08, linear warmup of 500 steps, and a decay factor of 0.1 at 70\% and 90\% of the training schedule. Total training takes about 1.2 days. For the ablations, we train for 25k steps (0.25x) on the same learning rate schedule. We set the learning rate of the pretrained image encoder and text encoder to be 10\% of the rest of the model which trains from scratch \cite{yang2020improving,deng2021transvg}.

We apply random scale jittering uniformly sampled between [0.4, 2.5] for every input image. The image is padded or randomly cropped to the size of (640, 640) after the scale jittering. For the ablation studies, we reduce the scale jittering magnitude to [0.8, 1.25] due to the shorter training. For detection and text-based localization tasks we also apply random horizontal flip following the standard protocol \cite{faster}. In addition, we tokenize the text with SentencePiece \cite{kudo2018sentencepiece} following T5 \cite{T5} and set the maximum expression length to 64 for all tasks.

\vspace{-3mm}
\section{Experiments}
\vspace{-1mm}

We compare FindIt to the state of the art (SOTA) on REC, LOC and DET tasks (Section~\ref{sec:main}). We follow the protocols established in prior works \cite{gupta2021gpv,yu2016modeling}, using only MS-COCO~\cite{mscoco} for training and validation. In addition, we evaluate how FindIt generalizes to OOD datasets and settings (Section~\ref{sec:ood}).

Here we define the family of FindIt models. \textit{FindIt} is trained jointly on REC, LOC, and DET tasks, while \textit{FindIt-REC, FindIt-LOC, FindIt-DET} are trained on each individual task to serve as single-task baselines. FindIt does not require more labeled data than existing REC methods, because pre-trained detector outputs \cite{yu2018mattnet,zhang2018grounding} or initialization with detector weights \cite{yang2019fast,yang2020improving,deng2021transvg} have been commonly used. Towards further unification, \textit{FindIt-MIX} trains on all RefCOCO splits (as opposed to a single RefCOCO split used by \textit{FindIt}), LOC, and DET together, resulting in one model for all splits instead of one model for each split, which is the case with FindIt. To our best knowledge, we are the first to report single-model unified training results on RefCOCO benchmarks.  We report all FindIt-MIX (384) results as an average of five independent runs.

\begin{table*}[t]
	\small
    \caption{Comparison with state-of-the-art methods on RefCOCO including those using external data for pretraining. We outperform the state of the art on RefCOCO~\cite{yu2016modeling}, RefCOCO+~\cite{yu2016modeling} and RefCOCOg~\cite{mao2016generation} with only R50 backbone. FindIt-REC is our own single task baseline. \textbf{Bold} indicates the highest non-unified training number. \textit{\first{Red}} indicates the highest number overall, whereas \textit{\second{blue}} the second highest. (Best viewed in color)}
    \label{tab:refcoco_results}
	\begin{center}
		\scalebox{0.65}[0.65]{
			\setlength
			\tabcolsep{8.4pt}
			\begin{tabular}{| c | c | c c c | c c c | c c c |}
				\hline
				\multirow{2}{*}{Models} & \multirow{2}{*}{Backbone} & \multicolumn{3}{c|}{RefCOCO} & \multicolumn{3}{c|}{RefCOCO+} & \multicolumn{3}{c|}{RefCOCOg} \\ 
				
				&  & val & testA & testB & val & testA & testB & val-g & val-u & test-u \\
				\hline \hline
				\textbf{\textit{Two-stage:}} & & & & & & & & & &\\
				CMN~\cite{hu2017modeling}  & VGG16 & - & 71.03 & 65.77 & - & 54.32 & 47.76 & 57.47 & - & - \\
				VC~\cite{zhang2018grounding}  & VGG16 & - & 73.33 & 67.44 & - & 58.40 & 53.18 &62.30 & - & - \\
				ParalAttn~\cite{zhuang2018parallel}  & VGG16 & - & 75.31 & 65.52 & - & 61.34 & 50.86 & 58.03 & - & - \\
				MAttNet~\cite{yu2018mattnet}  &	R101 & 76.65 & 81.14 & 69.99 & 65.33 & 71.62 & 56.02 & -  & 66.58 & 67.27 \\
				LGRANs~\cite{wang2019neighbourhood}  & VGG16 & - & 76.60 & 66.40 & - & 64.00 & 53.40 & 61.78 & - & - \\
				DGA~\cite{yang2019dynamic}  & VGG16 & - & 78.42 & 65.53 & - & 69.07 & 51.99 & - & - & 63.28 \\ 
				RvG-Tree~\cite{hong2019learning}  & R101 & 75.06 & 78.61 & 69.85 & 63.51 & 67.45 & 56.66 & - & 66.95 & 66.51 \\
				NMTree~\cite{liu2019learning}  & R101 & 76.41 & 81.21 & 70.09 & 66.46 & 72.02 &  57.52 & 64.62 & 65.87 & 66.44 \\
				CM-Att-Erase~\cite{liu2019improving} & R101 &	78.35 &	83.14 & 71.32 &	68.09 &	73.65 &	58.03 & - & 67.99 & 68.67 \\ 
				\hline
				\textbf{\textit{One-stage:}} & & & & & & & & & &\\
				SSG~\cite{chen2018real} &  DarkNet-53 & - & 76.51 & 67.50 & - & 62.14 & 49.27 & 47.47 & 58.80 & - \\  
				FAOA~\cite{yang2019fast} & DarkNet-53 & 72.54 & 74.35 & 68.50 & 56.81 & 60.23 & 49.60 & 56.12 & 61.33 & 60.36 \\
				RCCF~\cite{liao2020real} & DLA-34 & - & 81.06 & 71.85 & - & 70.35 & 56.32 & -  & - & 65.73 \\
				ReSC-Large~\cite{yang2020improving} & DarkNet-53 & 77.63 & 80.45 & 72.30 & 63.59 & 68.36 & 56.81 & 63.12 & 67.30 & 67.20 \\
				MCN~\cite{Luo2020MultiTaskCN} & DarkNet-53 & 80.08 & 82.29 & 74.98 & 67.16 & 72.86 & 57.31 &  - & 66.46 & 66.01 \\
				\hline
				\textbf{\textit{Transformer:}} & & & & & & & & & &\\
				TransVG \cite{deng2021transvg} & R50 & 80.32 & 82.67 & 78.12 & 63.50 & 68.15 & 55.63 & 66.56 & 67.66 & 67.44 \\
				TransVG \cite{deng2021transvg} & R101 & 81.02 & 82.72 & 78.35 & 64.82 & 70.70 & 56.94 & 67.02 & 68.67 & 67.73 \\
				FindIt-REC & R50 & 79.45 & 82.43 & 72.94 & 66.01 & 71.13 & 58.62 & 63.91 & 67.73 & 68.77 \\
				FindIt & R50 & \textbf{84.66} & \textbf{85.50} & \textbf{83.46} & \textbf{73.85} & \textbf{78.57} & \textbf{67.31} & \textbf{73.25}  & \textbf{77.64}  & \textbf{77.02} \\
				\hline
				\textbf{\textit{Unified Training:}} & & & & & & & & & &\\
				FindIt-MIX & R50 & 84.92 & 85.54 & 83.44 & 74.31 & 76.93 & 69.91 & 82.77 & 83.17 & 84.11\\
                FindIt-MIX (384) & R50 &  \second{87.09} & 85.55 & \second{86.89} & 76.35 & 75.47 & \second{71.85}	& \second{89.84} & \second{90.40} & \second{91.01} \\
                FindIt-MIX (384) & R101 &  \first{87.91} & 86.56 & \first{88.04} & \second{77.24} & 77.42 & \first{73.12} & \first{90.58} & \first{90.97} & \first{91.72} \\
				\hline
				\textbf{\textit{External Data:}} & & & & & & & & & &\\
				UNITER-L \cite{chen2020uniter} & R101 & 81.41 & 87.04 & 74.17 & 75.90 & 81.45 & 66.70 & - &  74.86 & 75.77 \\
				VILLA-L \cite{villa} & R101  & 82.39 & \second{87.48} & 74.84 & 76.17 & \second{81.54} & 66.84 & - & 76.18 & 76.71 \\
				MDETR \cite{mdetr} & R101 & 86.75 & \first{89.58} & 81.41 & \first{79.52} & \first{84.09} & 70.62 & - & 81.64 & 80.89 \\
				\hline
			\end{tabular}
		} 
		\vspace{-6mm}
	\end{center}
\end{table*}

\begin{table}[t]
\caption{Text-based Localization and Detection Benchmarks. All models in the tables use the ResNet-50 (R50) backbone.}
\label{tab:loc_and_det_results}
\begin{subtable}{0.48 \textwidth}
    \centering
    \caption{Text-based localization results on COCO. We compare with the single- and multi-task GPV \cite{gupta2021gpv}.}
    \label{tab:localization}
    \scalebox{0.8}{
    \begin{tabular}{|c|c|c|c|}
    \toprule
     Models  & Multitask & Image Size & AP50 \\
    \midrule
 FRCNN~\cite{faster,gupta2021gpv} & \xmark & 640 & 75.2 \\
 GPV~\cite{gupta2021gpv} & \cmark & 640 & 73.0\\
     \midrule
 FindIt-LOC  & \xmark & 640 & 77.9 \\
 FindIt-MIX & \cmark & 640 &78.6 \\
 FindIt & \cmark & 640 & \textbf{79.7 $\pm$ 0.1} \\
    \bottomrule
    \end{tabular}
    }
\end{subtable}
\hfill
\begin{subtable}{0.48 \textwidth}
    \centering
    \caption{Detection results on COCO. We compare with the single- and multi-task baselines from \cite{hu2021unit,Detectron2018}. .}
    \label{tab:detection}
    \scalebox{0.8}{ 
    \begin{tabular}{|c|c|c|c|}
    \toprule
     Models  & Multitask & Image Size & mAP  \\
    \midrule
 FRCNN \cite{Detectron2018} & \xmark & 1333 & 37.9 \\
 UNiT~\cite{hu2021unit} &  \xmark & 1333 & \textbf{40.6} \\
 UNiT~\cite{hu2021unit} &  \cmark & 1333 & 39.0\\
     \midrule
 FindIt-DET\footnotemark[1]  &  \xmark & 1024 & \textbf{40.6}\\
 FindIt-MIX & \cmark & 640 &  38.4\\
 FindIt & \cmark & 640 & 39.7 $\pm$ 0.1 \\
    \bottomrule
    \end{tabular}
    }
    \vspace{-4mm}
\end{subtable}
\end{table}

\vspace{-3mm}
\subsection{Main results}
\label{sec:main}

\autoref{tab:refcoco_results}, \autoref{tab:localization}, and \autoref{tab:detection} show our results on REC, LOC and DET tasks compared to the SOTA. In each table, we compare FindIt to both single- or multi-task approaches for the corresponding task. The single-task approaches are advantaged as they are fully optimized for the task. 

\autoref{tab:refcoco_results} compares with existing COCO-trained methods on the three popular REC benchmarks: RefCOCO~\cite{yu2016modeling}, RefCOCO+~\cite{yu2016modeling} and RefCOCOg~\cite{mao2016generation}. We see that FindIt outperforms the SOTA results, including two-stage/one-stage methods and recent Transformer-based models. In particular, on the challenging splits of RefCOCO+ (no location-based information) and RefCOCOg (longer expressions), FindIt outperforms the SOTA results by a clear margin of 5-10 points. Compared to the single-task baselines, FindIt consistently improves the performance by 3-9 points across the RefCOCO splits, showing the benefits of multitask training. 

We note that all results in \autoref{tab:refcoco_results} only use COCO box annotation and language corpus pretraining \cite{T5}. We do not pretrain on vision and language datasets or use the mask annotations in COCO \cite{Luo2020MultiTaskCN}. Existing approaches \cite{chen2020uniter,villa,mdetr} obtain SOTA performance on RefCOCO by pretraining on large vision and language datasets \cite{beer2021cc12m,im2text_sbu}, visual grounding datasets ~\cite{mdetr,krishnavisualgenome,plummer2015flickr30k}, or graph relationships~\cite{ernievil}. Without using external data, FindIt-MIX is on par with or better than the SOTA method \cite{mdetr} pre-trained with more visual grounding data. Our best-performing model on REC uses a smaller image size (384) than the rest of the paper (640).

To avoid contamination for FindIt, we remove the overlapping images of the RefCOCO val/test sets from the training sets of LOC and DET based on the RefCOCO split they are trained with. For FindIt-MIX, we carefully remove the overlapping images of all RefCOCO val/test sets from all REC, LOC, and DET training sets. The mixing ratio for FindIt-MIX is 2:2:1:1:1:1 among DET:LOC:REC:REC+:REC-g:REC-umd. The FindIt and FindIt-MIX models in \autoref{tab:refcoco_results} and \autoref{tab:loc_and_det_results} are the same without task-specific fine-tuning. 

\footnotetext[1]{FindIt-DET is trained and tested on COCO 17' to match the settings of~\cite{hu2021unit,Detectron2018}.}

\begin{figure*}[t]
    \centering
    \includegraphics[width=0.98\linewidth]{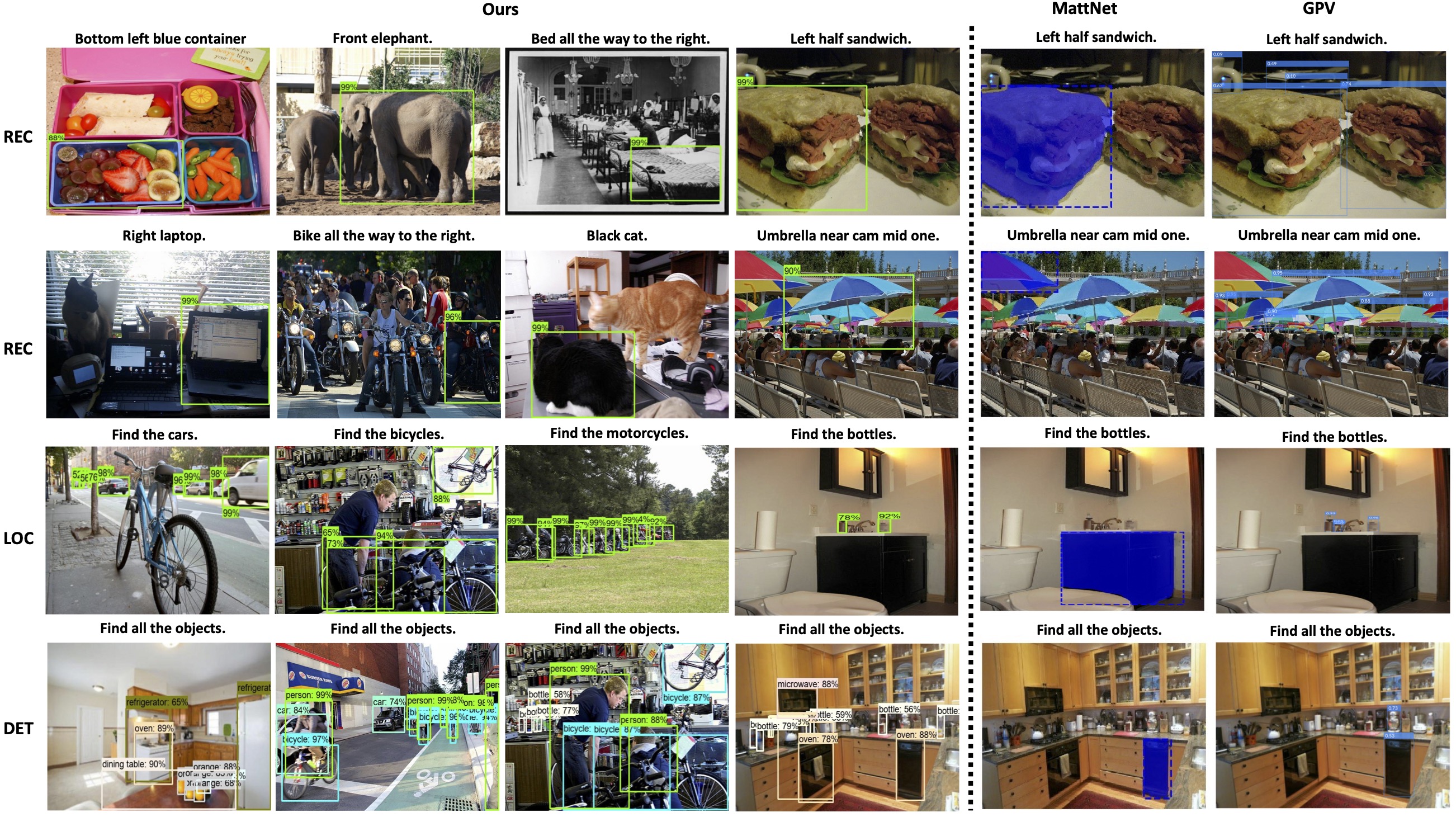}
    \caption{Visualization of FindIt on REC, LOC, and DET. Compared to existing baselines~\cite{yu2018mattnet,gupta2021gpv}, FindIt can perform these tasks well and in a single model.}
    \label{fig:multitask_vis}
    \vspace{-4mm}
\end{figure*}

\autoref{tab:localization} compares our work on the text-based localization (LOC) task. We compare to the recent GPV method~\cite{gupta2021gpv} which is is the best approach on this task. For FindIt, we report the mean and standard deviation over four individual RefCOCO splits. FindIt outperforms GPV in all settings. Following GPV \cite{gupta2021gpv}, we train both LOC and DET tasks on COCO'14 train split (80k images) and report performance on COCO'14 val split (40k images) in \autoref{tab:loc_and_det_results}. \autoref{tab:detection} shows our results on detection. We see that our approach is comparable to the full UNiT \cite{hu2021unit}, which uses a detection-specific task head, larger image size, and more training images (COCO'17 vs our COCO'14). Compared to the single-task setup, FindIt shows a similar performance gap to that seen in UNiT's multitask setup.  \autoref{fig:multitask_vis} shows examples of FindIt on all three tasks.

\begin{table}[t]
  \caption{Generalization study through text-based localization task.}
    \label{tab:gen}
    \begin{subtable}{0.5 \textwidth}
        \caption{Generalization to novel categories.}
        \label{tab:oos}     
        \centering
        \begin{tabular}{|l|c|ccc|}
        \toprule
         Model & REC & Base-80 & Novel-285 & All \\
        \midrule 
     FindIt-REC  & 79.5 & 21.3 & 5.2 & 13.1\\
     FindIt-LOC & - & 56.7 & 15.2 & 33.9 \\
     FindIt-MIX & \textbf{84.9} & \textbf{57.8} & \textbf{18.7} & \textbf{36.4} \\
        \bottomrule
        \end{tabular}
    \end{subtable}
    \hfill
    \begin{subtable}{0.45 \linewidth}
        \caption{Generalization to super-categories.}
        \label{tab:super_catg}
        \begin{tabular}{|l|ccc|}
        \toprule
         Model & COCO & COCO-O365 & O365 \\
       \midrule    
     FindIt-REC & 33.0 & 18.6 & 11.0 \\
     FindIt-LOC & 45.8 & 25.3 & 15.3 \\
     FindIt-MIX & \textbf{49.5} & \textbf{30.1} & \textbf{17.5} \\
        \bottomrule
        \end{tabular}
        \end{subtable}
\end{table}

\subsection{Generalization Capabilities of FindIt}
\label{sec:ood}

We now evaluate the generalization capabilities of the FindIt model presented in Section ~\ref{sec:main}. The Objects365 dataset~\cite{Objects365} is chosen for the study, because it is independently collected and represents OOD (Out-of-Distribution) data. In addition, the dataset is large, well-annotated with high recall, and contains all of 80 COCO categories and 285 novel categories (365 in total) to assess the generalization of FindIt models. Our models acquire the linguistic knowledge of novel categories from multi-task cross-attention learning and language pretraining \cite{T5}. However, as all of our single- and multi-task models share the same language pretraining, the main differences arise from multi-task learning.

\textbf{Localization on Novel Categories.} Even though referring expression models are able to effectively localize objects from complex queries, we want to investigate whether they are able to handle the text-based localization task. Thus, we evaluate the single-task FindIt-REC, FindIt-LOC, and unified training FindIt-MIX models on Objects365 dataset. All FindIt models are identical to their counterparts in \autoref{tab:refcoco_results} and \autoref{tab:localization} without further fine-tuning. The FindIt-REC model was trained on the RefCOCO UNC split. \autoref{tab:oos} shows the results, where the column ``Base-80'' evaluates the 80 COCO categories; ``Novel-285'' evaluates the 285 non-COCO categories; ``All'' evaluates all 365 categories; ``REC'' is the performance on RefCOCO UNC. We first observe that FindIt-REC struggles on this task , despite having strong performance on REC. FindIt-LOC model performs much better because it was directly trained for this task. Compared to FindIt-LOC and FindIt-REC, FindIt generalizes better especially on the novel categories of Objects365, because it has acquired broader knowledge about objects and grounding texts through multi-task learning.

\textbf{Localization on Super-categories.} By accepting text inputs, FindIt model relaxes the requirement for a pre-defined set of classes for localization and can generalize beyond the training vocabulary (i.e. COCO categories). We study this behavior by testing on COCO and Objects365 super-categories (e.g. giraffe $\in$ animal, pizza $\in$ food). The setup is identical to \autoref{tab:oos} except that the query category names are replaced with their corresponding super-categories. All models here are the same as in \autoref{tab:oos}. We present the results in ~\autoref{tab:super_catg}. The column ``COCO'' evaluates the COCO super-categories on COCO data; ``COCO-O365'' evaluates the COCO super-categories on Objects365; ``O365'' evaluates the Objects365 super-categories on Objects365. Despite the challenging setup, FindIt generalizes better than single-task baselines by a clear margin, showing the merits of broader grounding knowledge provided by multitask learning (see \autoref{fig:teaser} for more examples).

\begin{table}[t]
\centering
\caption{Runtime benchmark with recent REC approaches.}
\vspace{2mm}
\label{tab:speed_benchmark}
\resizebox{0.6\textwidth}{!}{
\begin{tabular}{| c | c | c | c | c |} 
 \hline
 Models & Image Size & Backbone & Runtime (ms) \\
 \hline \hline
 MattNet \cite{yu2018mattnet} & 1000 & R101 & 378 \\
 FAOA \cite{yang2019fast} & 256 & DarkNet-53 & 39 \\
 MCN \cite{Luo2020MultiTaskCN} & 416 & DarkNet-53 & 56 \\
 TransVG \cite{deng2021transvg} & 640 & R50 & 62 \\
 \hline
 FindIt & 640 & R50 & 107 \\
 FindIt & 384 & R50 & 57 \\
 \hline 
\end{tabular}
}
\vspace{-3mm}
\end{table}

\begin{table}[t]
    \caption{Ablations on task prompts, language model sizes, multi-level fusion architecture design, and mixing ratios.}
    \begin{subtable}{0.5 \linewidth}
    \caption{Ablations on task prompts. The first row corresponds to default FindIt. 
    }
    \label{tab:text_prompt}
    \centering
    \scalebox{0.95}{
    \begin{tabular}{| c | c | c | c |} 
     \hline
     LOC Prompt & DET Prompt & LOC & DET \\
     \hline \hline
     “Find the X” & “Find all the objects” & 78.78 & 38.96 \\
     \hline
     “X” & “This is detection task” & 79.13 & 38.88 \\
     \hline
     GPV \cite{gupta2021gpv} & “Find all the objects” & 78.92 & 38.97 \\
     \hline 
    \end{tabular}
    }
    \end{subtable}
    \hfill
    \begin{subtable}[h]{0.4 \linewidth}
    \caption{Ablations on language model sizes.}
    \label{tab:ablations_lm_size}
    \scalebox{0.9}{
    \begin{tabular}{|c|c|c|c|}
    \toprule
     Language Model & DET & LOC & REC \\
    \midrule
    T5-Small \cite{T5} & 38.7 & 78.7 & 80.7 \\
    T5-Base \cite{T5} & 38.4 & 78.7 & 81.0 \\
    T5-Large \cite{T5} & 38.8 & 78.8 & 81.2 \\
    \bottomrule
    \end{tabular}
    }
    \end{subtable} \\
    \begin{subtable}{0.4 \linewidth}
    \caption{Ablations on the fusion mechanism, feature dimension and the number of transformer layers.
    }
    \label{tab:ablations_fusion}
    \begin{tabular}{|c|c|c|c|c|c|}
    \toprule
     Fusion  & Dim. & \# Layers & DET & LOC & REC \\
    \midrule
 Concat  & 256 & 1 & 35.6 & 76.6 & 77.7 \\
 Product  & 256 & 1 & 35.4 & 76.6 & 78.9 \\
 Product  & 256 & 3 & 35.1 & 76.2 & 76.7 \\
    \midrule
 Attention  & 128 & 1 & 35.7 & 75.8 & 75.2 \\
 Attention  & 256 & 3 & 35.6 & 76.5 & 79.3 \\
 Attention  & 512 & 6 & 35.7 & 76.9 & 79.3 \\
 Attention  & 1024 & 12 & 35.7 & 77.1 & 82.1 \\
    \bottomrule
    \end{tabular}
    \end{subtable}
    \hfill
    \begin{subtable}{0.45 \linewidth}
    \centering
    \caption{Ablations on the fusion levels, feature dimension dimensionsion and the number of transformer layers.}
    \label{tab:ablations_fusion2}
    \begin{tabular}{|c|c|c|c|c|c|}
    \toprule
     Levels  & Dim. & \# Layers & DET & LOC & REC\\
    \midrule
 (5,) & 256 & 3 & 35.6 & 76.7 & 78.8 \\
 (5,) & 512 & 6 & 34.6 & 76.0 & 80.0 \\
 (5,) & 1024 & 12 & 33.0 & 75.0 & 80.5 \\  
 \midrule
 (4, 5) & 256 & 3 & 35.6 & 76.5 & 79.3 \\
 (4, 5) & 512 & 6 & 35.7 & 76.9 & 79.3 \\
 (4, 5) & 1024 & 12 & 35.7 & 77.1 & 82.1 \\
    \bottomrule
    \end{tabular}
    \end{subtable} \\
    \begin{subtable}[h]{0.5 \linewidth}
    \caption{Ablations on the fusion architecture for the REC tasks.}
    \label{tab:ref_all_ablations}
    \scalebox{0.8}{ 
    \begin{tabular}{|c|c|c|cccc|}
    \toprule
     Fusion & Dim. & Layers & UNC & Plus & G & UMD\\
    \midrule
     Concat.  & 128 & 1 & 76.1 & 60.4 & 53.2 & 62.4 \\ 
     Concat. & 256 & 3 & 76.8 & 61.7 & 54.5 & 63.6 \\  
     Concat. & 512 & 6 & 77.2 & 61.8 & 57.1 & 64.6 \\
    \midrule
     Product  & 128 & 1 & 66.5 & 62.4 & 55.3 & 63.3 \\ 
     Product  & 256 & 3 & 76.0 & 60.9 & 54.6 & 62.6 \\  
     Product  & 512 & 6 & 75.6 & 60.1 & 57.4 & 62.4 \\
     \midrule
     Attention  & 128 & 1 & 73.7 & 57.1 & 53.9 & 60.6 \\
     Attention  & 256 & 3 & 78.6 & 62.9 & 60.8 & 64.4 \\
     Attention  & 512 & 6 & 78.6 & 65.6 & 60.6 & 67.3 \\
    \bottomrule
    \end{tabular}
    }
    \end{subtable}
    \hfill
    \begin{subtable}[h]{0.45 \linewidth}
    \caption{Ablations on multitask mixing ratios for all tasks.}
    \label{tab:ablations_mixing2}
    \begin{tabular}{|c|c|c|c|}
    \toprule
     DET : LOC : REC & DET & LOC & REC \\
    \midrule
 1 : 1 : 1 & 35.5 & 76.6 & 78.5 \\
 2 : 1 : 1 & 35.9 & 75.9 & 77.7 \\
 1 : 2 : 1 & 34.9 & 77.0 & 78.3 \\
 2 : 2 : 1 & 35.8 & 76.8 & 76.9 \\
    \bottomrule
    \end{tabular}
    \end{subtable}
\end{table}

\subsection{Analysis and Ablations}
\label{sec:ablation}

\noindent {\bf Inference Time.} We benchmark the inference times across image sizes in \autoref{tab:speed_benchmark} on the REC task. FindIt is efficient and comparable with existing approaches, while achieving higher accuracy (See \autoref{tab:refcoco_results}). For fair comparison, all running times are measured on one GTX 1080Ti GPU. Compared to the two-step approach \cite{yu2018mattnet}, FindIt is more efficient because it trains end-to-end without a need for pre-computed detections.

\noindent {\bf Task Prompts.} We conducted the ablations on the prompts of LOC and DET tasks in ~\autoref{tab:text_prompt} and found the prompts have minimal effects on performance. Our LOC prompt ``Find the X'' is one of the prompts used by GPV \cite{gupta2021gpv}.

\noindent {\bf Language Model Size.} We conducted ablations on the language model sizes in ~\autoref{tab:ablations_lm_size} and found that larger models are only marginally better. In ~\autoref{tab:ablations_lm_size}, REC is the mean performance over all RefCOCO splits. We choose T5 base \cite{T5} as the best trade-off between performance and speed.  

\noindent {\bf Multi-level Fusion Architecture.} We conduct ablation studies on the fusion architecture and multitask mixing ratios. All experiments of this section are run with a 6x shorter schedule and weaker data augmentation for faster convergence. In all tables we use RefCOCO-UNC as a representative split to evaluate the REC task except for the ablation on language model size. In ~\autoref{tab:ablations_fusion} we study the effect of architecture choices on the downstream tasks. We find that attention-based fusion outperforms other alternatives given the same configuration (e.g. 256 dim, 3 layers). In addition, increasing the number of attention layers and the embedding dimension both improve the performance on referring expression, but not as much on detection and localization. We explore multi-scale fusion in ~\autoref{tab:ablations_fusion2}, and find that using more levels is beneficial for all model sizes we study. Thus, levels (4, 5) are chosen for all experiments. \autoref{tab:ref_all_ablations} delves deeper to show the benefits of attention fusion for REC tasks. With adequate model capacity (e.g. 256 dim, 3 layers), attention fusion outperforms the other alternatives under the same configuration. On the split with the most complex queries (RefCOCO-g), we notice attention fusion performs substantially better than other alternatives. From these studies, we choose (Attention, 256 dim, 3 layers) as our model, because we find the larger alternative (Attention, 512 dim, 6 layers) to perform only marginally better with full training schedule.

~\autoref{tab:ablations_mixing2} studies the sampling weight in multitask learning. We find that a simple 1:1:1 ratio achieves a good balance between DET, LOC and REC task performance. Increasing the sampling rate for one task tends to improve the performance at the expense of other tasks. We note that the mixing ratios can be further optimized to improve the performance of any constituent task. We use 256-dimension fusion features, 3 layers, and fusion levels (4, 5) in this ablation.

\vspace{-2mm}
\section{Conclusion}
We present Findit, which unifies referring expression comprehension, text-based localization, and object detection tasks. We propose multi-scale cross-attention to unify the disparate localization requirements of these tasks. Without any task-specific design, FindIt surpasses the state of the art on referring expression and text-based localization, shows competitive performance on detection, and generalizes better to out-of-distribution data and novel classes. All of these are accomplished in a single, unified and efficient model.\\

\noindent {\bf Acknowledgements.} We would like to thank Ashish Vaswani, Prajit Ramachandran, Niki Parmar, David Luan, Tsung-Yi Lin, and other colleagues at Google Research for their advice and helpful discussion.

\clearpage
%
%
\bibliographystyle{splncs04}
\bibliography{egbib}

\clearpage

\section*{Appendix A: FindIt Visualizations}
\begin{figure}[t]
    \centering
    \includegraphics[width=0.9\textwidth]{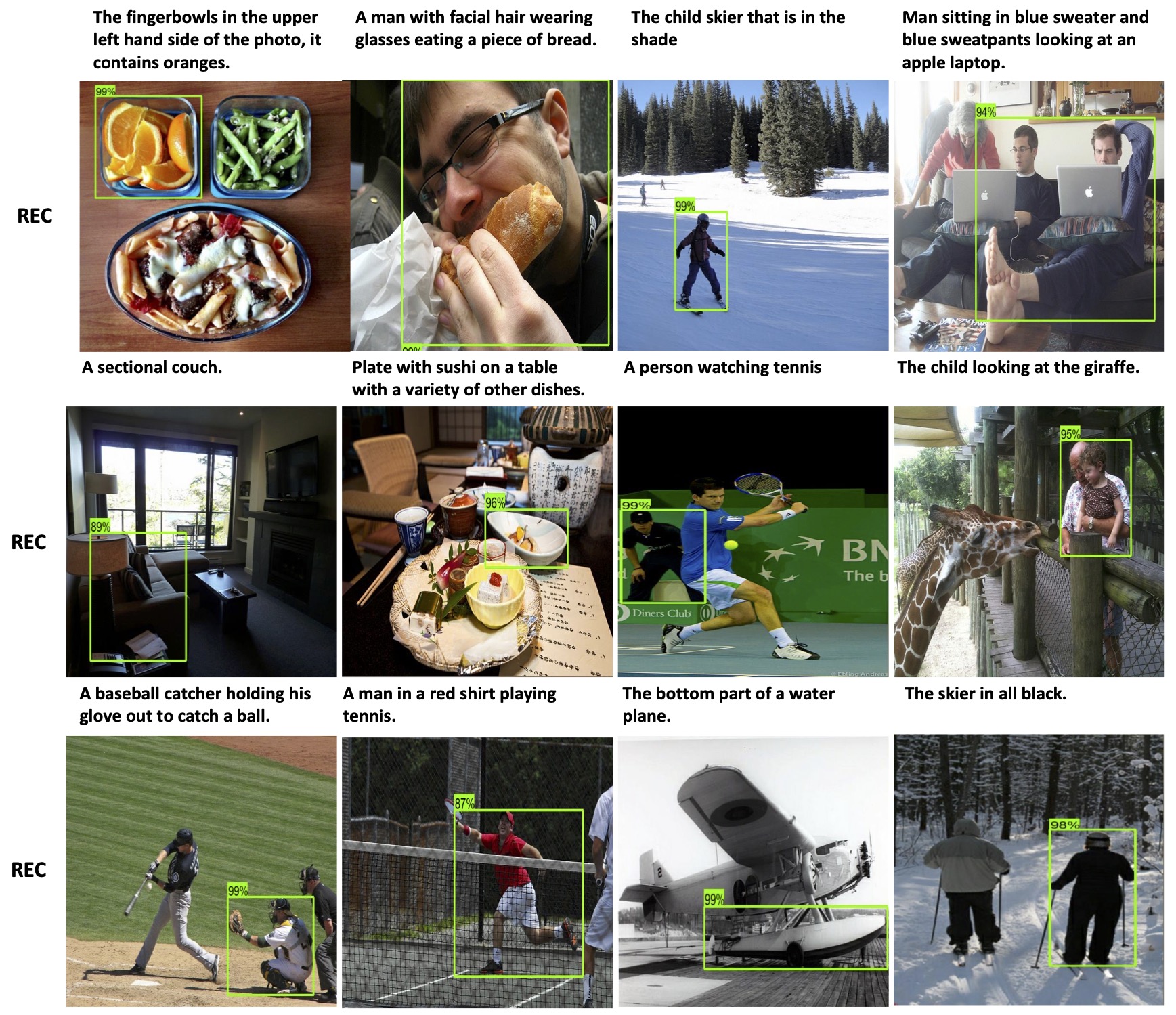}
    \includegraphics[width=0.9\textwidth]{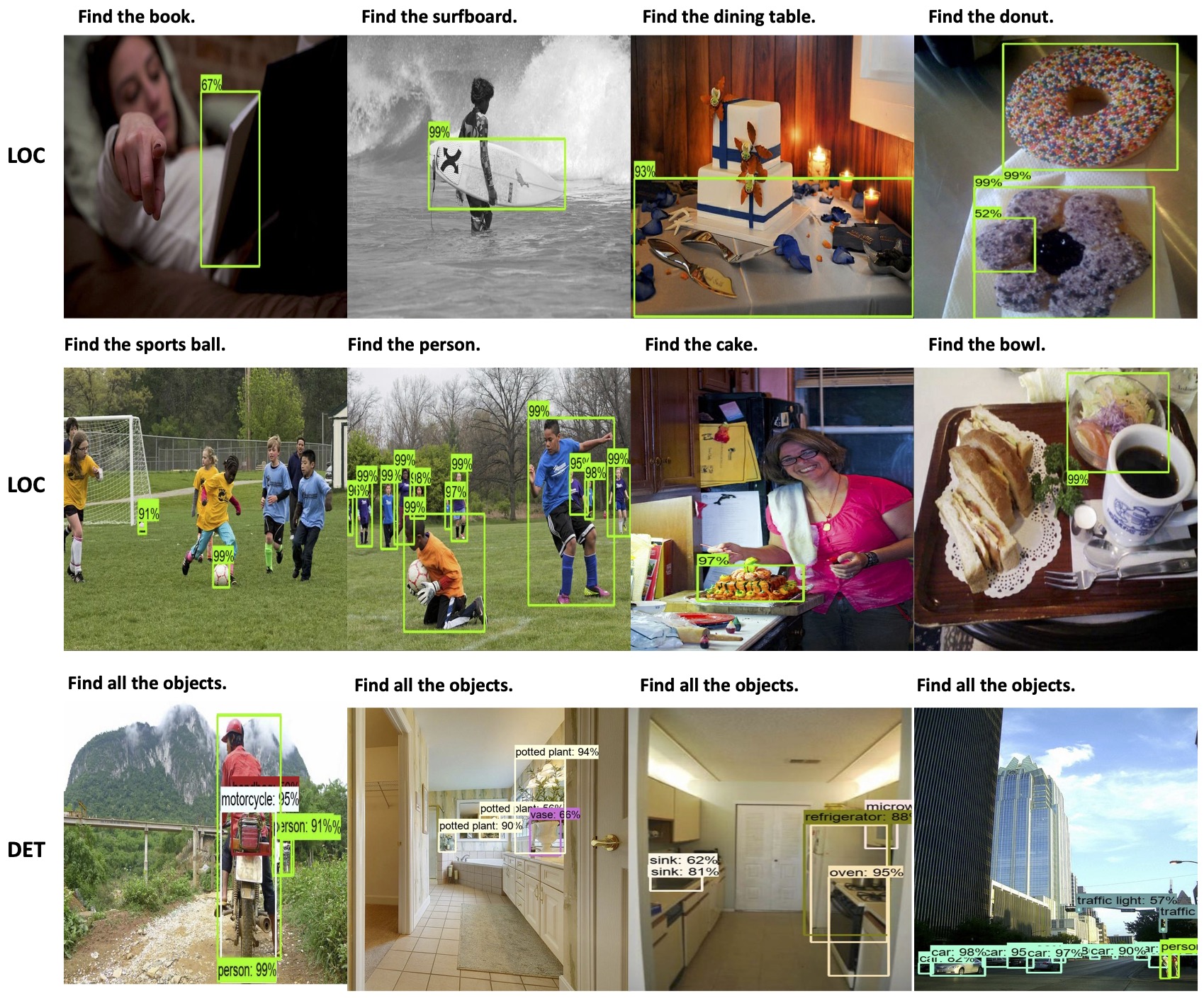}
    \caption{FindIt visualization on REC, LOC, and DET tasks (best viewed in color).}
    \label{fig:vis_supp}
\end{figure}

\autoref{fig:vis_supp} shows more visualizations of FindIt. The model is able to answer many complex expressions correctly and can even localize novel concepts such as ``bottom part of a water plane''. In addition, the model is able to localize many objects based on the category queries such as ``Find the sports ball'', and respond accurately to a detection query ``Find all the objects''.

\vspace{-1mm}
\section*{Appendix B: Failure Mode Analysis}

\begin{figure}[t]
    \centering
    \includegraphics[width=0.45\linewidth]{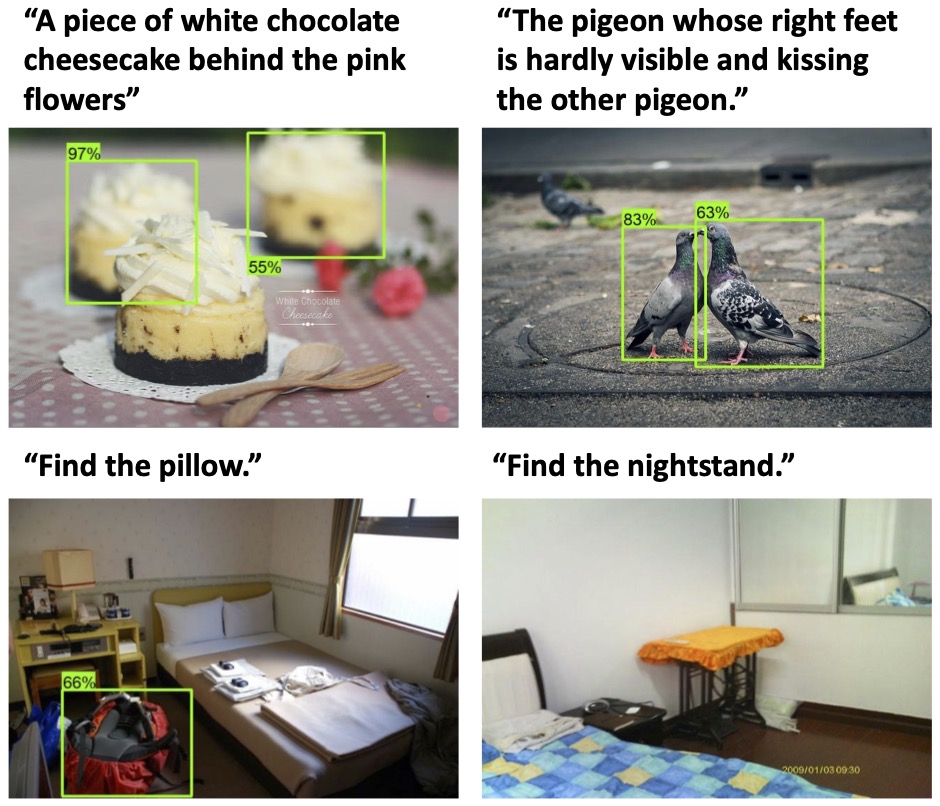}
    \includegraphics[width=0.45\linewidth]{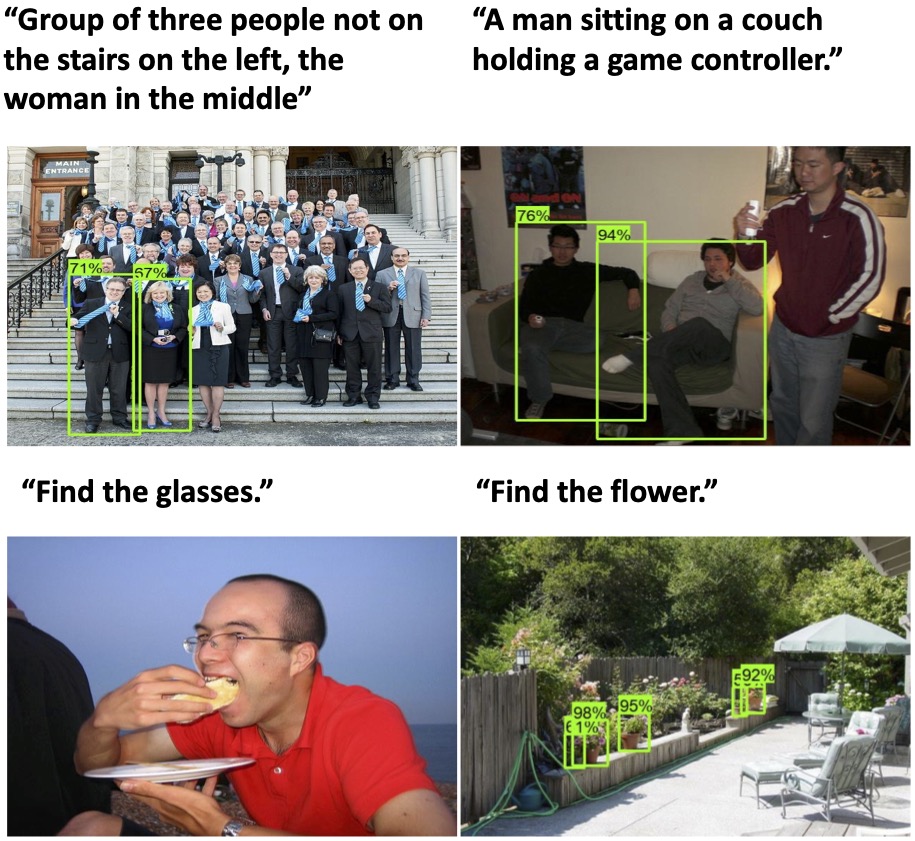}
    \caption{Failure cases of FindIt on the referring expression comprehension and the text-based localization tasks (best viewed in color).}
    \label{fig:vis_failure}
\end{figure}

In \autoref{fig:vis_failure}, we visualize the failure cases of FindIt. The model tends to struggle with confounding objects of similar categories or attributes especially when given a complex query. The rare, long-tail and novel categories also pose challenges. These can be remedied by training on referring expression datasets with confounding objects \cite{chen2020copsref}, larger detection datasets \cite{Objects365}, or pretraining on large language and vision datasets~\cite{beer2021cc12m,clip,align}.

\vspace{-1mm}
\section*{Appendix C: Mixture Ablations}

\autoref{tab:ablations_full_mixing} studies the mixing ratio of detection and localization with all RefCOCO splits together. 
Each row in ~\autoref{tab:ablations_full_mixing} is a single unified model able to perform REC, LOC, and DET tasks on all RefCOCO splits.  Unlike the other ablations, we compare the mixtures here on the full training schedule. We observe that increasing the detection and localization mixing ratio tends to improve the detection and localization tasks, but at a slight cost to the referring expressions comprehension tasks.

\begin{table*}[h]
\caption{Ablations on FindIt mixtures with all RefCOCO splits. The mixture is given as the ratios between Det:Loc:Ref:Ref+:Ref-g:Ref-umd tasks.}
	\small
	\begin{center}
		\scalebox{0.7}[0.7]{
			\setlength
			\tabcolsep{8.4pt}
			\begin{tabular}{| c | c | c | c c c | c c c | c c c |}
				\hline
				\multirow{2}{*}{Mixture} & \multirow{2}{*}{DET} & \multirow{2}{*}{LOC} & \multicolumn{3}{c|}{RefCOCO} & \multicolumn{3}{c|}{RefCOCO+} & \multicolumn{3}{c|}{RefCOCOg} \\ 
				
				& &  & val & testA & testB & val & testA & testB & val-g & val-u & test-u \\
				\hline \hline
				2:2:1:1:1:1 & 38.4 & 78.7 & 84.92 & 85.54 & 83.44 & 74.31  & 76.93 & 69.91 & 82.77 & 83.17  & 84.11 \\
				3:3:1:1:1:1 & 39.3 & 79.3 & 84.27 & 85.52 & 83.80 & 74.56 & 76.06 & 68.75 & 82.56 & 83.09 & 83.58  \\
				4:4:1:1:1:1 & 39.7 & 79.7 & 83.69 & 83.45 & 82.99 & 72.63 & 73.68 & 68.64 & 80.72 & 81.03 & 81.75  \\
				\hline
			\end{tabular}
		} 
	\end{center}
\label{tab:ablations_full_mixing}
\end{table*}

\clearpage

\end{document}